# More-or-Less CP-Networks

**Fusun Yaman** and **Marie desJardins**

Department of Electrical Engineering and Computer Science
University of Maryland Baltimore County
Baltimore,MD
{ fusun, mariedj }@cs.umbc.edu

## Abstract

Preferences play an important role in our everyday lives. CP-networks, or CP-nets in short, are graphical models for representing conditional qualitative preferences under *ceteris paribus* ("all else being equal") assumptions. Despite their intuitive nature and rich representation, dominance testing with CP-nets is computationally complex, even when the CP-nets are restricted to binary-valued preferences. Tractable algorithms exist for binary CP-nets, but these algorithms are incomplete for multi-valued CP-nets. In this paper, we identify a class of multi-valued CP-nets, which we call *more-or-less CP-nets*, that have the same computational complexity as binary CP-nets. More-or-less CP-nets exploit the monotonicity of the attribute values and use intervals to aggregate values that induce similar preferences. We then present a search control rule for dominance testing that effectively prunes the search space while preserving completeness.

## 1 Introduction

Humans and businesses often need to choose between different ways to achieve certain goals; in general, the choice depends on some set of preferences. These preferences may have conditional relationships. For example, you might like to see Paris more than Orlando during your vacation, but if you have children with you, then you might prefer Orlando, since it offers entertainment for the entire family. Such preferences can be represented in a *conditional preference network* (CP-net) [4].

CP-nets are intuitive graphical models for representing conditional qualitative preferences over a set of outcomes, which are represented as attribute vectors. Inference tasks in CP-nets include outcome optimization (determining the most preferred of all possible outcomes), dominance queries (determining whether one outcome is strictly preferred to another, under *all* interpretations of the CP-net), and ordering queries (depending whether one outcome may be preferred to another, under *some* interpretation of the CP-net). In general, optimization and ordering queries are easy—even in multi-valued CP-nets—but dominance queries are hard. For example, Boutilier et al. [4] present an algorithm for dominance testing for binary, tree-structured CP-nets, which has quadratic complexity; however, this algorithm is shown to be incomplete for multi-valued tree-structured CP-nets.

One approach for reducing the number of values is to group multiple values. For example, we might divide business hours into two intervals: morning (8 am to noon) and afternoon (noon to 5 pm), thereby reducing the number of values from nine to two. However, with this approach we also lose some expressivity. For example, if we prefer earlier hours to later hours, then we can only say that morning is better than afternoon. We are unable to express that of the morning hours, 8 am is better than 11 am This issue is also recognized in qualitative decision theory and referred as the *resolution* problem [10].

In many cases, we may be able to take advantage of properties of the relative preferences over the values of certain variables. In particular, ordinal variables are a common special case of multi-valued variables where we often observe (1) monotonicity of preferences over the attribute values and (2) conditional preferences associated with the variable's children that are the same across a range of values.

In this paper, we identify a class of multi-valued CP-nets, which we call *more-or-less CP-nets*, that has the same computational complexity as binary CP-nets. More-or-less CP-nets take advantage of common properties of preferences over ordinal variables, by exploiting monotonicity of the preferences over the attribute values, and by using intervals to aggregate ranges of values that induce similar preferences. This representation is the first step towards our long-term goal of learning CP-nets from observations in complex domains.

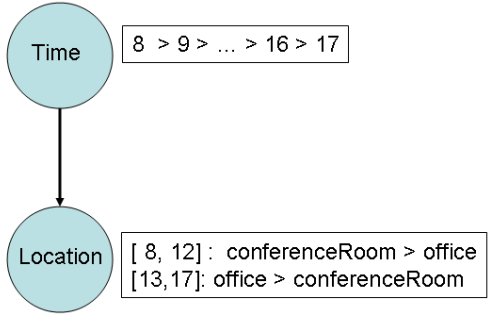

Figure 1: Meeting time and location preferences.

Figure 1 shows an example of a more-or-less CP-net that expresses the "earlier is better" preference: earlier meetings are preferred to later meetings (*monotonicity*); furthermore, if the meeting is in the morning, then I prefer meeting in the conference room, whereas if it is in the afternoon I prefer meeting in my office (*aggregation*).

In the following sections, we briefly give some background on CP-nets, and then present more-or-less CP-nets. Next, we present modifications to dominance testing algorithms for more-or-less CP-nets, based on existing algorithms for binary CP-nets, that have the same computational complexity as the binary versions. After discussing related work, we conclude with future research directions.

## 2    Background on CP-Nets

A CP-net is a compact and natural representation of preference statements that uses a graphical representation to express knowledge about conditional dependence and independence under a *ceteris paribus* (all else being equal) interpretation. In this section we summarize the model and semantics of CP-Nets as presented by Boutilier *et al.* [4, 5].

Given a set of variables $V = \{X_1, X_2 \ldots X_n\}$ whose domains are $Dom(X_1) \ldots D(X_n)$, an outcome $o$ is an assignment of variables in $V$ such that every $X_i \in V$ is mapped to a value in $Dom(X_i)$. Referring to the set of all outcomes as $O$, a preference ranking is a total preorder $\succeq$ over $O$, where $o_1 \succ o_2$ means that $o_1$ is strictly preferred to $o_2$, and $o_1 \succeq o_2$ means that outcome $o_1$ is equal or preferred to $o_2$.

A partial assignment $x$ to $X \subset V$ maps each variable in $X$ to a value in its domain. If $X$ and $Y$ are disjoint subsets of $V$ and $x$ and $y$ are assignments to $X$ and $Y$ then $xy$ is an assignment to $X \cup Y$.

For any $X \subseteq V$ and $Y = V - X$, $X$ is said to be *preferentially independent* of $Y$ iff for every pair of assignments $x_1, x_2$ to $X$ and $y_1, y_2$ to $Y$, the following statement holds:

$$x_1 y_1 \succeq x_2 y_1 \Leftrightarrow x_1 y_2 \succeq x_2 y_2. \tag{1}$$

That is, the values of the variables in $Y$ have no effect on the relative preference over the joint assignment of variables in $X$. Suppose that $X$, $Y$ and $Z$ are non-empty partitions of $V$. Then $X$ is said to be *conditionally preferentially independent* of $Y$, given an assignment $z$ to $Z$, iff for every pair of assignments $x_1, x_2$ to $X$ and $y_1, y_2$ to $Y$, the following statement holds:

$$x_1 y_1 z \succeq x_2 y_1 z \Leftrightarrow x_1 y_2 z \succeq x_2 y_2 z. \tag{2}$$

In this case, as long as we hold the values of the $Z$ variables fixed, the $Y$ variables have no effect on the preference over the variables in $X$.

**Definition 1 (CP-net)** *A CP-net over variables $V = \{X_1, \ldots X_n\}$ is a directed graph $G$ over $X_1, \ldots X_n$ whose nodes are annotated with conditional preference tables $CPT(X_i)$ for each $X_i \in V$.*

Using $Pa(X)$ to refer to the parents of node $X$ in the graph $G$, the semantics of satisfying a CP-net is given by the following definition:

**Definition 2 (Satisfying a CP-net)** *Let $N$ be a CP-net over variables V, $X_i \in V$ be some variable, and U be $Pa(X_i)$, the parents of $X_i$ in N. Let $Y = V - \{U \cup X_i\}$. Let $\succeq_u^i$ be the ordering over $Dom(X_i)$ dictated by $CPT(X_i)$ for a particular instantiation u of $X_i$'s parents. Suppose that $\succ$ is a preference ranking over every possible assignment of V.*

*A preference ranking $\succ$ satisfies $\succeq_u^i$ iff for all assignments $y_1, y_2$ to Y and for all $x_1, x_2 \in Dom(X_i)$, $y_1 x_1 u \succ y_2 x_2 u$ whenever $x_1 \succeq_u^i x_2$.*

*A preference ranking $\succ$ satisfies $CPT(X_i)$ if it satisfies $\succeq_u^i$ for every assignment u to U. A preference ranking $\succ$ satisfies the CP-net N iff it satisfies $CPT(X_i)$ for each variable $X_i$.*

Intuitively, every CP-net induces a partial order on the possible set of outcomes; we refer to this induced partial order as the *induced preference graph*. A preference ranking satisfies a CP-net if and only if it is a topological sort of the CP-net's induced preference graph. We also say that such a preference ranking is *consistent* with the CP-net. In general, there can be multiple topological sorts, so there can be multiple consistent preference rankings for a given CP-net.

Any acyclic CP-net is satisfiable and from this point on we base our discussions on acyclic CP-nets only.

Given a CP-net, there are three reasoning tasks of interest:

- Outcome Optimization: Given a partially instantiated outcome, find the assignment for the rest of the variables that is maximally preferred. Boutilier *et al.* [5] proposed a forward sweep algorithm to compute the optimal completion of a partially specified outcome. The forward sweep algorithm runs in $O(n)$

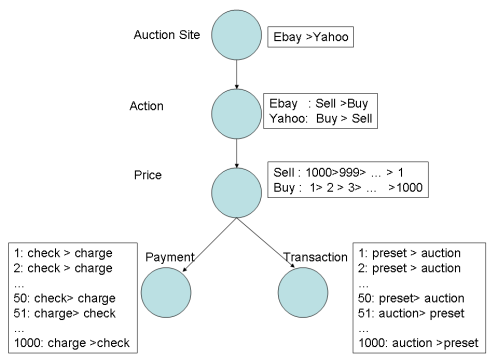

Figure 2: Preferences of an auctioneer.

time, where $n$ is the number of attributes in the CP-net.

- Ordering Queries: Order a set of outcomes such that the ordering is consistent with the CP-net. As shown by Boutilier *et al.* [5], we can test whether an outcome $o$ can be ordered before an outcome $p$ in $O(n)$ time. Furthermore, we can order $m$ outcomes consistently with a CP-net in $O(nm^2)$ time.

- Dominance Queries: Determine whether an outcome $o$ is always preferred to another outcome $p$ (i.e., whether it is strictly preferred under *all* consistent preference rankings). This is the most complex reasoning task. For binary tree-structured CP-nets, Boutilier *et al.* [5] present an $O(n^2)$ algorithm. For singly directed-path connected, binary CP-nets, dominance testing is NP-complete. It has not been established whether dominance testing for multiply connected, binary CP-nets is in NP.

## 3 More-or-Less CP-nets

Consider the CP-net in Figure 2, which represents the preferences of an online auctioneer. It states that the auctioneer prefers e-Bay auctions to Yahoo auctions in general. Also, he prefers to sell items at e-Bay, but prefers to buy items at a Yahoo auction. Naturally, he prefers the higher price to a lower price when he is selling items, and the other way around when he is buying something. Furthermore, if the item costs less than or equal to $50, then he prefers the payment to be made by a personal check, but if the cost is more than $50, then he thinks a credit card payment is more secure. Finally, if the item costs less than $50, he is willing to skip the auction and buy or sell directly at the asking price, but if the item costs more than $50, he wants to take his chances with the auction.

Although CP-nets can adequately represent this simple example, it is not a scalable representation. It would be impractical, for example, to answer dominance queries, with

1000 possible values to consider for the Price variable. Also note that the CPTs for the dependents of the Price variable has 1000 entries. Since the size of the CPT grows exponentially with the number of the parents of a node, memory limitations will also arise in complex domains, with many nodes and multiple parents.

In this example, however, the Price variable exhibits a monotonic preference structure, the directionality of which depends on the parent value (Action). Also, consecutive entries in the Payment and Transaction CPTs are always the same, except when Price=50. Thus we can aggregate the table entries as: $1 \leq price \leq 50 : check > charge$ and $51 \leq price \leq 1000 : charge > check$ for the Payment CPT. In effect, we create two categories for the Price attribute: *less than or equal to 50* and *more than 50*. Although we utilize these categories in the CPT tables of its children, we do not restrict the Price variable to be binary.

We now generalize this property of the $Price$ attribute to define the *monotonic variables* of a CP-net.

**Definition 3 (Monotonic Variable)** *Let $N$ be a CP-net over variables $V$. A variable $X \in V$ is a monotonic variable of $N$ iff there exists a total order $\sqsubset$ on $Dom(X)$ and a value $c \in Dom(X)$ such that two constraints are satisfied:*

- Monotonicity constraint: *Every preference ranking in $CPT(X)$ is either the same order as $\sqsubset$ induces on $Dom(X)$ or the reverse.*

- More-or-less constraint: *If there is an edge from $X$ to $Y$ in $N$, and if two assignments to $Pa(Y)$, $p_1$ and $p_2$, differ only at the value of $X$, then the entries in $CPT(Y)$ for $p_1$ and $p_2$ are the same whenever:*

  – *$p_1(X) \sqsubseteq c$ and $p_2(X) \sqsubseteq c$ or*
  – *$c \sqsubset p_1(X)$ and $c \sqsubset p_2(X)$,*

  *where $p_i(X)$ is the value of $X$ in $Pa(X)$.*

*Furthermore, the two categories of $X$ are the sets $less(X) = \{x \mid x \in Dom(X) \wedge x \sqsubseteq c\}$ and $more(X) = Dom(X) - less(X)$.*

The monotonicity constraint asserts that the conditional preferences over the values of $X$ are either monotonically increasing or monotonically decreasing. The more-or-less constraint asserts that there is some value $c$ of $X$ that serves as a "break point" for the preferences of $X$'s children. That is, for values up to $c$ (i.e., the $less(X)$ values), $X$'s children exhibit a fixed behavior, and for values above $c$ ($more(X)$), they exhibit a different fixed behavior.

By this definition, $Price$ is a monotonic variable, where every value less than or equal to 50 is in $less(X)$, and every value above 50 is in $more(X)$. Note that according to this definition, every binary variable in a CP-net is also a monotonic variable. We now define more-or-less CP-nets, which are a subset of CP-nets with the restriction that all of the variables must be monotonic.

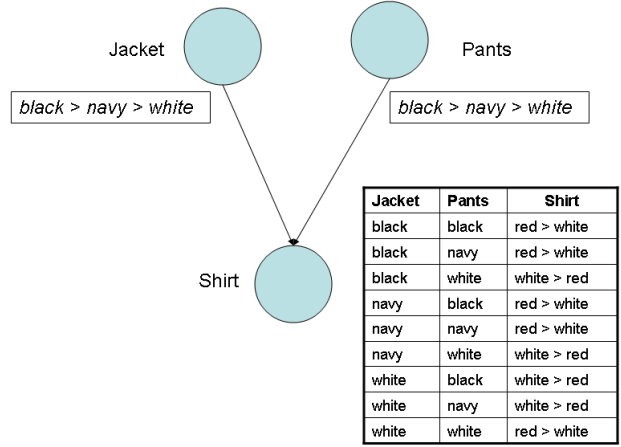

Figure 3: Evening dress.

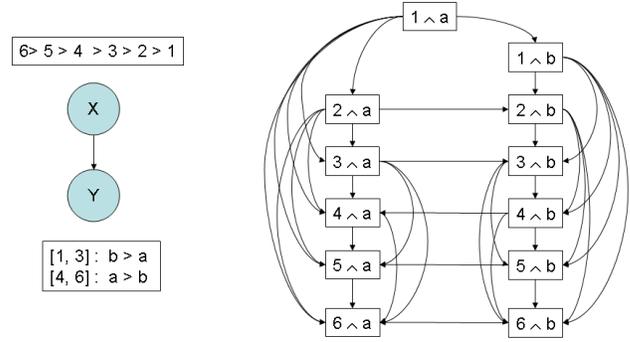

Figure 4: A more-or-less CP-net and the preference graph it induces.

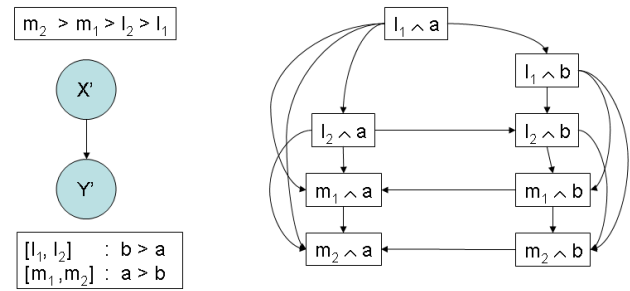

Figure 5: The compact version of the more-or-less CP-net in Figure 4.

**Definition 4 (More-or-Less CP-net)** *A CP-net $N$ over variables $V$ is a more-or-less CP-net iff every variable in $V$ is a monotonic variable in $N$.*

Any binary CP-net is a more-or-less CP-net, because binary variables are always monotonic.

The meeting preference in Figure 1 and the auction preferences in Figure 2 are both more-or-less CP-nets. Figure 3 is another example of a more-or-less CP-net that shows the evening dress preferences[1] of a person whose shirt color depends the color of his pants and jacket. He unconditionally prefers black to navy to white for both the pants and the jacket. If both his pants and jacket are dark colored (black or navy) or both of them are white then he prefers a red shirt. Otherwise he prefers a white shirt. Note that the conditional preference table of the shirt color satisfies the more-or-less constraint for both the jacket and pant colors.

The semantics of more-or-less CP-nets are the same as given in Definition 2.

## 4 Reasoning with More-or-Less CP-Nets

Existing outcome optimization and ordering algorithms for CP-nets are effective for binary CP-nets as well as multi-valued ones. Unfortunately, this is not the case for dominance queries. Boutilier *et al.* [5] showed that answering a dominance query corresponds to finding a directed path from one outcome to the other in the induced preference graph. In CP-nets with multi-valued variables, this search space (i.e., the number of nodes in the induced preference graph) grows exponentially. However, CP-nets with monotonic variables result in preference graphs with a specific structure that we can exploit to achieve better performance. In particular, we can prune some of the nodes in the preference graph of a more-or-less CP-net. The following example illustrates this point.

**Example 1** *Figure 4 shows a more-or-less CP-net $N$ and its induced preference graph. In this preference graph, every outcome that differs only at $X$ is totally ordered. Figure 5 shows another more-or-less CP-net $N'$ and its induced preference graph. Note the similarity between $N'$ and $N$. Even though $X'$ has only four values, we can map any outcome pair $o$ and $p$ of $N$ to an outcome pair $o'$ and $p'$ in $N'$ such that $o$ dominates $p$ in $N$ iff $o'$ dominates $p'$ in $N'$. It is easy to see that this would still hold for an arbitrary domain size of $X$.*

Every path from an outcome $o$ to $p$ in the induced preference graph coincides with an *improving flipping sequence* $o = o_1; o_2, \ldots, o_{k-1}, o_k = p$, where every $o_i$ differs from $o_{i+1}$ only at one variable $(X)$, and $o_{i+1}$ assigns a more preferred value to $X$ than $o_i$. An *irreducible flipping sequence* is an improving sequence such that, when a subsequence of it is removed, it is no longer an improving flipping sequence. Intuitively, irreducible flipping sequences limit unnecessary changes to a variable. Unfortunately, multi-valued more-or-less CP-nets allow irreducible flipping sequences with redundant flippings; therefore, not all irreducible flipping sequences are equally efficient.

---

[1]This is an extension of the example in [5]. The original CP-net contains only binary valued variables.

Once again, consider the CP-net in Figure 4. Note that $1a$, $2a$, $2b$, $5b$ is an irreducible flipping sequence from $1a$ to $5b$, because if we remove any subsequence, it is no longer an improving flipping sequence. However, a more intuitive and efficient improving flipping sequence would be: $1a$, $1b$, $5b$. This sequence skips the flip from 1 to 2 and then 2 to 5, by changing the value of X from 1 directly to 5.

We now define *representative sets* and *skip-flipping* sequences with respect to two outcomes.

**Definition 5 (Representative Set)** *Let $N$ be a more-or-less CP-net over variables $V$. Suppose that $X$ is a variable in $V$. $R = \{x_1, x_2\}$, a subset of $Dom(X)$, is a representative set of $X$ iff $x_1$ and $x_2$ are in different categories of X, i.e., either $x_1 \in more(X)$ and $x_2 \in less(X)$, or vice versa.*

The representative set $R$ consists of two representative values from the domain of $X$ that include one value from each category. For example, in the CP-net in Figure 4, $\{2, 4\}$ and $\{1, 5\}$ are representative sets of $X$.

**Definition 6 (Skip-Flipping Sequence)** *Let $N$ be a more-or-less CP-net over variables $V$ and $F = o_1 o_2 \ldots o_k$ be an irreducible flipping sequence of $N$. Suppose that for every variable $X \in V$, $Rep(X, o_1, o_k)$ is a representative set of $X$ such that $o_k(X) \in Rep(X, o_1, o_k)$ and $o_1(X) \in Rep(X, o_1, o_k)$ iff $o_1(X)$ and $o_k(X)$ are in different categories of X or $o_1(X) = o_k(X)$. Then $F$ is a skip-flipping sequence if, for every variable $X$ and every $o_i$ that flips the value of $X$, $o_i(X) \in Rep(X, o_1, o_2)$.*

**Example 2** *Let $N$ be the more-or-less CP-net in Figure 4 and $F$ be the irreducible flipping sequence $o$=1a; 1b;p=5b. Then $Rep(X, o, p) = \{1, 5\}$, $Rep(Y, o, p) = \{a, b\}$, and $F$ is a skip-flipping sequence (because every flip of X and Y results a value in $Rep(X, o, p)$ and $Rep(Y, o, p)$). On the other hand, o=1a, 2a, 2b, p = 5b is not a skip-flipping sequence, because it flips the value of X from 1 to 2, which is not in $Rep(X, o, p)$.*

*Consider the flipping sequence $F''$ where o=4b, 4a,p=6a. For this sequence, $Rep(X, o_1, o_k)$ can be any of the sets $\{1, 6\}$, $\{2, 6\}$, or $\{3, 6\}$, because $4 \in more(X)$ and $6 \in more(X)$. Note that, regardless of which of these sets are chosen, $F''$ is a skip-flipping sequence.*

In general, while searching for an improving flipping sequence from $o$ to $p$, instead of trying every improving flip for a variable $X$, we limit the flipping values to the elements of $Rep(X, o, p)$. In this way, we can avoid noncritical flipping. We call this the *critical-flipping rule*. The following lemma states that the critical-flipping rule does not compromise completeness.

**Lemma 1** *Let $N$ be a more-or-less CP-net and $o$ and $p$ be two outcomes. If there is an irreducible flipping sequence from $o$ to $p$, then there is a skip-flipping sequence with respect to $o$ and $p$.*

Proof Sketch: Given an irreducible flipping sequence $F = o_1, \ldots, o_k$, we can construct a skip-flipping sequence. For simplicity, assume that for every variable $X$ in $N$, $o_1(X) \in less(X)$ and $o_k(X) \in more(X)$. First, replace every $o_i(X) \in less(X)$ with $o_1(x)$, and every $o_i(X) \in more(X)$ with $o_k(x)$. The resulting sequence might not be an improving sequence, since it might have $o_i = o_{i+1}$ for some $i$. However, it is easy to see that $o_{i+1}$ is never worse than $o_i$. Therefore, if we eliminate these repetitions, we end up with an irreducible flipping sequence that only flips the variables to a representative value. The case where $o_1(X)$ and $o_k(X)$ are in the same set requires a slightly different value assignment step, but can be proven in a similar way.

**Theorem 1** *Let $N'$ be a more-or-less CP-net and $N$ be a binary CP-net. Suppose that answering dominance queries for $N$ is in complexity class $C$. If $N$ and $N'$ have similar graph structure (i.e. both are trees,or singly connected or multiply connected graphs), then answering dominance queries for $N'$ is also in complexity class $C$.*

Proof Sketch: Essentially, more-or-less CP-nets behave very much like binary CP-nets. Using the critical-flipping rule, we can avoid exploring non-critical portions of the search space. Furthermore, the critical-flipping rule ensures that we only consider flipping between two values. Next, we show that we can modify a dominance testing algorithm for a binary valued CP-net $N$ to work with a more-or-less CP-net $N'$ that has the same structure of $N$.

Algorithm 1 presents `abstractDT`, an abstract algorithm for answering a dominance query $N \models p \succ o$ for a binary valued CP-net $N$. Basically the algorithm searches for an improving flipping sequence from $o$ to $o'$. The function $select(N, o, o^*, o')$ returns a variable $X$ in $N$ such that if $p$ is the outcome obtained by flipping the value of $X$ in $o^*$, then $o^*p$ is an improving flip and there might be an improving flipping sequence from $p$ to $o'$. Note that $select(N, o, o^*, o')$ can be nondeterministic depending on the structure of $N$. Furthermore $select(N, o, o^*, o')$ prunes the search space, using some of the search control rules described by Boutilier *et al.* [5] such that completeness of the search is not effected. For example if N is tree structured then $select(N, o, o^*, o')$ applies the rules *suffix fixing, least variable flipping* and *forward pruning*.[2] Thus if

---

[2]The *least variable flipping rule* is known to be complete for binary valued tree-structured cp-nets. As a result it can not be used in $select(N, o, o^*, o')$ if N is not a tree.

---

**Algorithm 1** abstractDT$(N, o', o)$

---

/* An abstract dominance testing algorithm for
  binary CP-nets */
Set $o^* = o$
**while** $o^* \neq o'$ **do**
    $X = select(o, o^*, o')$
    **if** $X$ is empty **then**
        **return** false
    **end if**
    $flip(o^*, X)$
**end while**
**return** true

---

**Algorithm 2** ml-abstractDT$(N, o', o)$

---

/* An abstract dominance testing algorithm for
  more-or-less CP-nets */
**for each** variable $X$ **do**
    Let $R(X, o', o)$ be the representative set w.r.t. $o$ and $o'$
**end for**
Set $o^* = o$
**while** $o^* \neq o'$ **do**
    **for each** variable $X$ s.t. $o^*(X) \notin R(X, o', o)$ **do**
        $flipInCategory(o^*, X, R(X, o', o))$
    **end for**
    **if** $o^* \neq o'$ **then**
        **return** true
    **end if**
    $X = ml\text{-}select(N, o, o^*, o')$
    **if** $X$ is empty **then**
        **return** false
    **end if**
    $flipOutCategory(o^*, X, R(X, o', o))$
**end while**
**return** true

---

$select(N, o, o^*, o')$ returns no variable then there is no flipping sequence from $o$ to $o'$ that contains $o^*$.

Consider Algorithm 2, which is a modification of Algorithm 1. The function $flipInCategory(o^*, X, R(X, o', o))$ flips the value of X to a value $x \in R(X, o', o)$ iff it is an improving flip and $o^*(X)$ and $x$ are in the same category of $X$. Similarly, $flipOutCategory(o^*, X, R(X, o', o))$ flips the value of X from $o^*(X)$ to a value $x \in R(X, o', o)$ iff it is an improving flip and $o^*(X)$ and $x$ are in different categories of $X$. $ml\text{-}select(N, o, o^*, o')$ uses exactly the same algorithm as $select(N, o, o^*, o')$, except that for any variable $X$, flipping the value of $X$ means changing it to a value that is in another category of $X$. It is easy to see that the algorithm `ml-abstractDT` generates skip-flipping sequences. Furthermore, the complexity added by the additional step $flipInCategory(o^*, X, R(X, o', o))$ is $O(n)$ where $n$ is the number of variables in $N$. Note that for every $X$, the value of $o^*(X)$ is going to be flipped by $flipInCategory(o^*, X, R(X, o', o))$ at most once because once it is flipped to a value in $R(X, o', o)$ the condition $o^*(X) \notin R(X, o', o)$ would fail the next time[3]. Thus, the overall complexity of `ml-abstractDT` is in the same class as the complexity of `abstractDT`.

## 5   More Complex Preferences

More-or-less CP-nets enforce two constraints on every variable of a CP-net; monotonicity and a single critical point where the preference rankings change. However more-or-less CP-nets might still be able to represent preferences that do not satisfy these two constraints. For example consider a financial advisor who prefers to meet with his large-account clients over the lunch (around noon) and prefers meeting other clients early morning or late afternoon. If the meeting starts sometime between 11:00am and 1:00pm, he prefers to meet at a restaurant otherwise he meets at his office. The CP-net, N, in Figure 6a shows these preferences where $a \sim b$ means $a$ and $b$ are equally preferred. Note that the $meetingTime$ variable in $N$ does not satisfy the monotonicity constraint because the preference ranking on possible meeting times is not a total order. Also there are two critical points (11:00am and 1:00pm) where the meeting location preference changes based on the meeting time. Thus $meetingTime$ is not a monotonic variable and $N$ is not a CP-net.

However, if we encode the problem differently then we can to represent the preferences in $N$ as a more-or-less CP-net. Specifically, instead of $meetingTime$ we can use $D = \|meetingTime - 12 : 00pm\|$; distance from noon as a variable then we will have a monotonic variable. For large-account clients, the preference of $D$ will $0 > 1 > 2 > 3 > 4$ and for others $4 > 3 > 2 > 1 > 0$. Furthermore, if $D \leq 1$ then the preferred location is a restaurant and if $D > 1$ then it is the office. Figure 6b shows the more-or-less CP-net that represents the same preferences as the CP-net in Figure 6a.

## 6   Related Work

In decision theory representation of qualitative preferences have been studied [8]. Among those it is argued by Doyle *et al.*[9] that conditional preferences under the *ceteris paribus* assumption is a natural way to represent preferences. CP-nets is a graphical representation of conditional preferences evaluated with all else being equal assumption. CP-nets can represent the preferences in a natural and intuitive way but in general it is hard to reason with them. As shown by Lang *et al.* [11] the complexity of reasoning with CP-nets

---

[3]If $o(X)$ and $o'(X)$ were in different classes of $X$ then $R(X, o', o)$ contains $o(X)$ and $o'(X)$ and $flipInCategory(o^*, X, R(X, o', o))$ is never executed.

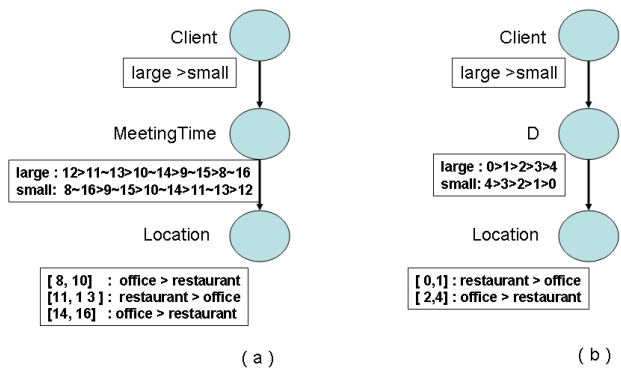

(a)                                                              (b)

Figure 6: Two different encodings for meeting preferences of a financial advisor. (a) is not a more-or-less CP-net because $meetingTime$ is not a monotonic variable. (b) is a more-or-less CP-net that has the variable $D$ representing the distance of the meeting time from noon.

is PSPACE-complete. For acyclic CP-nets the problem is NP-hard [6].

In an effort to deal with the complexity of reasoning, previous work focuses on approximating the partial order induced by a CP-net. To this end two different approaches were proposed based on UCP-nets [3] and soft constraints [7].UCP-nets are combination of additive models [1] and CP-nets. UCP-nets have a directed graph representing the conditional relationships however instead of preference rankings CPT tables contain quantitative utility values. Domshlack *et al.* [7] combines soft constraints [2] with CP-nets. In this approach, given an acyclic CP-net, a corresponding soft constraint satisfaction problem (SCSP) is constructed. The solution to the SCSP is guaranteed to be information preserving and satisfies the conditional properties of the underlying CP-net.

Our approach is orthogonal to previous work. In particular we concentrate on identifying a subclass of multi-valued CP-nets that has the same computational characteristics of binary valued CP-nets. The definition of more-or-less CP-nets ensures that every binary valued CP-net is also a more-or-less CP-net. In more-or-less CP-nets, we utilize the monotonicity of the variables and assume the existence of a single critical point per variable where the preference rankings change behavior. Although this approach is similar to the categorization notion used in qualitative decision theory [10] to model quantitative values qualitatively, it is not exactly the same. This is due to the fact that we allow the multi-valued attribute to attain values from its original domain (not just one value representing each category) and represent a preference ranking over the original domain with the monotonicity restriction. Thus our approach does not suffer from the resolution problem [10] which is an issue for most qualitative reasoning systems that represents quantitative values qualitatively.

## 7   Conclusions and Future Work

In this paper, we introduced *more-or-less CP-nets*, which are a special case of multi-valued CP-nets for ordinal variables. The variables in more-or-less CP-nets exhibit monotonic preferences and the "more-or-less constraint," which allows the preferences of a range of values to be aggregated together. We showed that efficient dominance-testing algorithm exists for more-or-less CP-nets.

In future work, our focus is on learning CP-nets from observed behavior (i.e., when a user chooses one outcome from a group of alternatives). This learning problem is a challenging one that has not been addressed by previous work. In this setting, learning more-or-less CP-nets will entail identifying the direction of monotonicity and the value of $C$ for monotonic variables. We also plan to generalize to non-monotonic ordinal preferences (e.g., "peak" values) and to aggregations that may be non-binary (i.e., with more than a single "break point"). The present paper only deals with more-or-less CP-nets, in which all of the variables are monotonic. In general, however, we may have CP-nets with different types of variables, which will require efficient hybrid inference methods.

## Acknowledgements

This work was supported by DARPA's Integrated Learning Program.